\pdfoutput=1
\documentclass[11pt]{article}

\usepackage{ACL2023}
\usepackage{tabularx}
\usepackage{longtable}
\usepackage{multirow}
\usepackage{array}
\newcommand{\PreserveBackslash}[1]{\let\temp=\\#1\let\\=\temp}
\newcolumntype{C}[1]{>{\PreserveBackslash\centering}p{#1}}
\newcolumntype{R}[1]{>{\PreserveBackslash\raggedleft}p{#1}}
\newcolumntype{L}[1]{>{\PreserveBackslash\raggedright}p{#1}}
\usepackage[T1]{fontenc}
\usepackage{times}
\title{PhayaThaiBERT: Enhancing a Pretrained Thai Language Model with Unassimilated Loanwords}

\author{
    Panyut Sriwirote \\
    Department of Linguistics \\
    Chulalongkorn University \\
    \texttt{panyutsriwirote@gmail.com} \And
    Jalinee Thapiang \\
    ClickNext \\
    \texttt{jalinee.t@clicknext.com} \AND
    Vasan Timtong \\
    ClickNext \\
    \texttt{vasan.t@clicknext.com} \And
    Attapol T. Rutherford \\
    Department of Linguistics \\
    Chulalongkorn University \\
    \texttt{attapol.t@chula.ac.th}
}

\begin{document}
\maketitle
\begin{abstract}

Although WangchanBERTa has become the de facto standard in transformer-based Thai language modeling, it still has shortcomings in regard to the understanding of foreign words, most notably English words, which are often borrowed without orthographic assimilation into Thai in many contexts. We identify the lack of foreign vocabulary in WangchanBERTa's tokenizer as the main source of these shortcomings. We then expand WangchanBERTa's vocabulary via vocabulary transfer from XLM-R's pretrained tokenizer and pretrain a new model using the expanded tokenizer, starting from WangchanBERTa's checkpoint, on a new dataset that is larger than the one used to train WangchanBERTa. Our results show that our new pretrained model, PhayaThaiBERT, outperforms WangchanBERTa in many downstream tasks and datasets.

\end{abstract}

\section{Introduction}

The use of pretrained transformer-based masked language models has led to state-of-the-art performance in many NLP tasks, both in English \cite{devlin2019bert} \cite{liu2019roberta} and other languages \cite{conneau-etal-2020-unsupervised} \cite{martin-etal-2020-camembert} \cite{nguyen-tuan-nguyen-2020-phobert}. The transformer-based architecture and its word embedding vocabulary are pretrained on a large amount of data on certain objective functions. And then we can perform supervised fine-tuning on the pretrained model for specific tasks. For Thai, WangchanBERTa, the state-of-the-art transformer-based language model, resulted in superior performance in many Thai NLP tasks \cite{lowphansirikul2021wangchanberta}, such as misspelling correction \cite{thai_misspelling_correction}, text classification \cite{thai_question_classification}, essay quality check \cite{thai_essay_checking}, and sentiment analysis \cite{thai_sentiment_analysis}.

Borrowing the architecture from RoBERTa \cite{liu2019roberta}, WangchanBERTa employs some special techniques aimed to address some characteristics of the Thai language, such as forcing its tokenizer to preserve spaces by replacing them with a special token that will never be included with other tokens, and normalizing sequences of more than 3 repeated characters, often used for emphasis in casual texts, to just 1 character. However, a characteristic of the Thai language underemphasized by this model is the code-switching nature of the language. Thai natural user-generated content usually contains orthographically unassimilated loanwords, foreign words that are written using non-Thai characters exactly like they would be in the original language. This phenomenon is most evident for English words, especially in discussions involving foreign culture or in academic context, where many important concepts simply do not have a widely accepted Thai equivalent. %Below is an example of such loanwords in Thai.\footnote{https://www.beartai.com/lifestyle/1309020}

%\begin{displayquote}
%\textthai{ผลจากการ Prompt คำสั่งให้ Midjourney คอย Generate ออกมา}

%`The result of prompting Midjourney to generate [some images]'
%\end{displayquote}

The previous model does not contain enough foreign vocabulary to address this nature. We hypothesize that the vocabulary of the transformer-based model should be expanded and pretrained to improve the performance of the downstream supervised fine-tuning process. 

The main aim of this study is, therefore, to train a new transformer-based masked Thai language model that has been enhanced to better understand instances of code-swtiched language and unassimilated loanwords. A secondary aim is to pretrain the model using even larger training data than the one used to train WangchanBERTa. The new pretrained model should then be able to achieve better performance. Indeed, our results in Section \ref{sec:fine_tuning_result} have shown this to be the case.

Our new pretrained model, which we call PhayaThaiBERT, is publicly available on Hugging Face Model Hub.\footnote{https://huggingface.co/clicknext/phayathaibert} The source code used in our experiment is available on our GitHub repository.\footnote{https://github.com/clicknext-ai/phayathaibert}

\section{Related Work}

Modern language modeling has largely been focused on transformer-based models as pioneered by the BERT architecture \cite{devlin2019bert}. The original BERT is implemented using 2 objectives, masked token prediction and next sentence prediction. Its tokenizer uses the WordPiece algorithm \cite{wu2016googles}. Later studies further improve upon this architecture and method, most notably the RoBERTa architecture \cite{liu2019roberta}, which drops the next sentence prediction objective and optimizes many aspect of BERT pretraining. New tokenization algorithms have since been proposed, most notably the unigram algorithm \cite{kudo-2018-subword}, which is reported to result in better performance compared to previously used tokenization algorithms. In addition to single-language modeling, multilingual models have also been implemented, most notably XLM-R \cite{conneau-etal-2020-unsupervised}, which combines the method proposed by RoBERTa with the unigram tokenization algorithm, along with other techniques to scale the model to 100 languages. It takes advantage of the SentencePiece library \cite{kudo-richardson-2018-sentencepiece}, which implements BPE and unigram tokenization algorithm in a way that is highly portable and language-agnostic. For Thai, a model has also been implemented. As stated in the introduction, WangchanBERTa, a RoBERTa-based model that uses SentencePiece's unigram tokenization, has been the de facto standard in Thai langauge modeling since its release in 2021. It is precisely this model that we wish to improve further.

\section{New Dataset and Vocabulary Expansion}

In order to achieve our goal of creating a language model capable of understanding unassimilated loanwords, we employ 2 strategies, 1) further pretraining on a new, larger dataset, and 2) expansion of WangchanBERTa's vocabulary to include foreign words.

\paragraph{New Dataset}
We create a large raw text dataset using various sources, including a large portion of the data used to train WangchanBERTa, most notably wisesight-large and pantip-large, which account for over 90\% of WangchanBERTa's training set. The inclusion of WangchanBERTa's data is in the hope that our model, which uses WangchanBERTa's weights as the starting point (see Section \ref{sec:start_checkpoint}), could have a chance to recalibrate its parameters to account for new tokenization resulting from added vocabulary. The total size of our preprocessed and tokenized training data is 156.5GB, 2 times larger than the one used to train WangchanBERTa. The list of sources we use to create our dataset is shown in the appendix.

\paragraph{Vocabulary Expansion}
One source of the shortcomings of WangchanBERTa when it comes to understanding unassimilated loanwords is its tokenizer, whose vocabulary contains only a small number of foreign words, making it unable to tokenize unassimilated words in a way that makes sense. In our experiment, we try improving it by adding new foreign vocabulary to WangchanBERTa's SentencePiece unigram tokenizer. This is done through a simple transfer of vocabulary, along with each word's unigram score, from XLM-R's tokenizer. Sine both tokenizers use SentencePiece's implementation of the unigram algorithm and both store their vocabulary in the Protobuf format\footnote{https://protobuf.dev/}, it is trivially easy to transfer vocabulary between them. We only transfer vocabulary that does not already exist in WangchanBERTa's vocabulary and we intentionally do not transfer XLM-R's vocabulary that consists of at least 1 character from the Thai Unicode block to avoid interfering with the tokenization of existing vocabulary. In addition to this, we also add all emojis as defined by the Unicode Consortium to the vocabulary. These emojis have no associated unigram scores and will be tokenized using simple pattern matching. We believe this is justifiable since we do not expect emojis to form any morphological or syntactic unit larger than 1 character.

After vocabulary expansion, the model's vocabulary size increases from 25,005 to 249,262, significantly increasing the size of the model from 106M parameters to 278M parameters. %Below is a comparison between tokenization done by WangchanBERTa's tokenizer and our expanded tokenizer.

%\begin{displayquote}
%WangchanBERTa:\newline
%\textthai{<s> | \_ | ผลจากการ | <\_> | \_ | pro | mp | t | <\_> | \_ | คําสั่งให้ | <\_> | \_ | mid | j | our | ne | y | <\_> | \_ | คอย | <\_> | \_ | gen | e | rate | <\_> | \_ | ออกมา | </s>}

%PhayaThaiBERT:\newline
%\textthai{<s> | \_ | ผลจากการ | <\_> | \_prompt | <\_> | \_ | คําสั่งให้ | <\_> | \_mid | jour | ney | <\_> | \_ | คอย | <\_> | \_generate | <\_> | \_ | ออกมา | </s>}
%\end{displayquote}

\section{Experimental Setup}
\label{sec:experiment}

\subsection{Model Architecture}

We implement the same model architecture as that of WangchanBERTa, namely the RoBERTa architecture, with some minor modification. Our architecture has 2 separate embedding lookup tables. One contains the embeddings of the existing vocabulary while the other contains the embeddings of the added vocabulary. During forward pass, the embedding of each token will be looked up from their respective tables. The result will then be combined in such a way as to be identical to the case with only 1 embedding lookup table. This ensures that there is no need to modify any of the other layers of the RoBERTa architecture. The reason for this modification will be discussed in section \ref{sec:trainingprocedure}.

The final model that we publish on Hugging Face Model Hub, however, does not use the architecture described above. It has been converted back to the RoBERTa architecture for compatibility and ease-of-use.

\subsection{Starting Model Checkpoint}
\label{sec:start_checkpoint}

We use the best checkpoint of WangchanBERTa (wangchanberta-base-att-spm-uncased) as the model's starting point. Every parameter weight of the model is transfered directly from said checkpoint, with the exception of the added vocabulary's embeddings, which are randomly initialized.

\subsection{Preprocessing and Tokenization}
\label{sec:preprocessing}

Since most of the model's parameters are transfered directly from WangchanBERTa, it is imperative that we also preprocess and tokenize our training data in the same manner. Accordingly, we preprocess and tokenize all of our training data following the procedure described in the Methodology section of the WangchanBERTa paper \cite{lowphansirikul2021wangchanberta}, with the exception of the tokenization section, where we will use our expanded tokenizer instead.

The resulting preprocessed and tokenized texts are then grouped into chunks, each no longer than 416 tokens. Pieces of texts that are by itself longer than 416 tokens are simply discarded. The final preprocessed and tokenized data is stored in the Arrow format\footnote{https://arrow.apache.org/} using Hugging Face's \textit{datasets} library \cite{lhoest-etal-2021-datasets}.

\subsection{Train-Validation-Test Splits}

After preprocessing and tokenization, our training data has a size of 91,130,926 examples (156.5GB). We then randomly split the data into 91,030,926 examples (156.3GB) for Train set, 50,000 examples (87.9MB) for Validation set, and 50,000 (88MB) examples for Test set.

\subsection{Pretraining Task}

The objective of the model's pretraining is masked language modeling. In the same manner as WangchanBERTa, we masked tokens at the subword level rather than the whole word. We sample 15\% of the tokens in a sequence as tokens to be masked. Masking is done by replacing the token with the special token <mask> 80\% of the time, replacing them with a random token from the vocabulary 10\% of the time, and leaving the token unchanged 10\% of the time. This is done on the fly as the training progresses, making the training samples different from epoch to epoch. The goal of the model is to predict what token is being masked by <mask>. Loss is computed using cross entropy over the entire vocabulary.

\subsection{Training Procedure}
\label{sec:trainingprocedure}

We discover that mere further pretraining from WangchanBERTa's checkpoint without any modification to the training procedure invariably results in a spike in validation loss after only a few hundred steps. We solve this problem by following some of the techniques proposed in \cite{howard-ruder-2018-universal}, which are:

\paragraph{Discriminative fine-tuning}
We use different learning rates for each layer in the model. We group all the model parameters into 15 layers, 12 of which correspond to the 12 transformer blocks while the remaining 3 correspond to 1) the embeddings of the added vocabulary, 2) the embeddings of the existing vocabulary, and 3) the masked language modeling head. The learning rate for the added vocabulary's embeddings are set to be the largest, as determined by peak learning rate (see appendix). All other layers' learning rates are set to be progressively lower the further they are from the last layer in the forward pass. Each layer's learning rate is set to be 2.6 times smaller than its next layer. ($lr^{l-1} = lr^l / 2.6$). Following this scheme, the layer with the smallest learning rate is the embeddings of the existing vocabulary.

\paragraph{Gradual unfreezing}
At the start of the training, we freeze every model parameter except those belonging to the embeddings of the added vocabulary. Other layers will then be gradually unfrozen every 1,000 steps. Among the yet-unfrozen layers, the one closest to the last layer in the forward pass will be chosen to be unfrozen at each unfreezing step. The grouping of parameters into layers is identical to the one used for discriminative fine-tuning, meaning the last layer to be unfrozen at step 14,000 is the embeddings of the existing vocabulary.

\paragraph{}
During the course of the training, we save a checkpoint of the model every 100 steps, and evaluate the model at that checkpoint with the Validation set. The computed loss is then used to decide whether the model has already converged or not. Training is done on the LANTA HPC cluster\footnote{https://thaisc.io/thaisc-resorces/lanta} using the Accelerator API of Hugging Face's \textit{accelerate} library \cite{accelerate}.

Due to resource and time limitation, we are able to train only one model in a single run, forcing us to train the model in a somewhat irregular manner. The irregular actions we are forced to perform are as follows:

\paragraph{Disabling discriminative fine-tuning}
We discover that the model fails to improve further at around step 4,200. We hypothesize that this is caused by our implementation of discriminative fine-tuning, which causes higher layers to overfit more quickly while lower layers still fail to converge due to their learning rate being too small. We then decide to disable discriminative fine-tuning, after which the model starts to improve again.

\paragraph{Periodically resetting the learning rate scheduler}
Since we cannot predict how many steps it would take for the model to converge, we cannot decide exactly what value we should set for \textit{max steps}. We decide to set the learning rate scheduler's max steps to 500,000, which, due to the way Hugging Face's \textit{accelerate} library handles multi-GPU training, will be reached within 31,250 actual update steps when trained on 16 GPUs. The scheduler will then be periodically reset when we subjectively decide that there is still potential for the model to improve using a larger learning rate.

\paragraph{Switching self-attention layers from FP16 to FP32 training}
At around step 6,200, the training becomes numerically unstable, with NaN consistently appearing in both training and evaluation to the point where the model cannot learn anything further. We determine that NaN always appears for the first time in the computation graph in self-attention layers. We then decide to switch those layers from FP16 to FP32 training, after which the problem is fixed for good.

\subsection{Evaluation}

In order to assess its capability, we fine-tune the pretrained model on several downstream tasks and datasets. We also fine-tune other similar models for comparison.

\subsubsection{Datasets}

The datasets are chosen as to be mostly similar to those in the WangchanBERTa paper \cite{lowphansirikul2021wangchanberta}. We use the train-validation-test split exactly as provided by the datasets hosted on Hugging Face Dataset Hub.\footnote{https://huggingface.co/datasets} In cases where certain splits are unavailable, we randomly split the provided train split using the \textit{train\_test\_split} method from Hugging Face's \textit{datasets} library \cite{lhoest-etal-2021-datasets} with seed = 2020, the same seed used in the WangchanBERTa paper. A summary of all the datasets used in the final evaluation is shown in Table \ref{tab:datasets}.

\begin{table*}
    \centering
    \footnotesize
    \begin{tabular}{llllll}
        \hline
        Name & Classification Task & Labels & Train & Validation & Test \\
        \hline
        1. wisesight\_sentiment & single-label, sequence & 4 & 21,628 & 2,404 & 2,671 \\
        2. wongnai\_reviews & single-label, sequence & 5 & 36,000\textsuperscript{*} & 4,000\textsuperscript{*} & 6,203 \\
        3. yelp\_review\_full\textsuperscript{**} & single-label, sequence & 5 & 36,000\textsuperscript{*} & 4,000\textsuperscript{*} & 6,200 \\
        4. generated\_reviews\_enth & single-label, sequence & 5 & 141,369 & 15,708 & 17,453 \\
        5. prachathai67k & multi-label, sequence & 12 & 54,379 & 6,721 & 6,789 \\
        6. thainer (ner) & single-label, token & 13\textsuperscript{***} & 5,079\textsuperscript{*} & 635\textsuperscript{*} & 634\textsuperscript{*} \\
        7. lst20 (pos) & single-label, token & 16 & 63,310 & 5,620 & 5,250 \\
        8. lst20 (ner) & single-label, token & 10 & 63,310 & 5,620 & 5,250 \\
        9. thai\_nner (layer 1) & single-label, token & 104 & 2,934\textsuperscript{*} & 980\textsuperscript{*} & 980 \\
        \hline
        \\
        \multicolumn{6}{l}{\textsuperscript{*} Randomly split from the provided train split with seed = 2020} \\
        \multicolumn{6}{l}{\textsuperscript{**} Downsampled from the full dataset} \\
        \multicolumn{6}{l}{\textsuperscript{***} The unconfirmed tag is removed}
    \end{tabular}
    \caption{Description of the datasets used in the final evaluation.}
    \label{tab:datasets}
\end{table*}

\paragraph{Wisesight Sentiment \cite{bact_2019_3457447}}
A dataset of 26,703 Thai messages from Thai social media. Each message is labeled as either \textit{positive}, \textit{neutral}, \textit{negative}, or \textit{question}.

\paragraph{Wongnai Reviews \cite{wongnai_reviews}}
A dataset of 46,203 Thai reviews from Wongnai, a Thai reviewing platform. Each review is labeled with a rating from 1 to 5 stars.

\paragraph{Yelp Review Full \cite{NIPS2015_250cf8b5}}
A dataset of 700,000 English reviews from Yelp, an American reviewing platform. Each review is labeled with a rating from 1 to 5 stars. We do not use the entire dataset for our fine-tuning. Instead, we split the provided train split into 2 new splits, train and validation, each with 600,000 and 50,000 reviews respectively. We then downsample all splits by taking only the front of each split. The exact number taken in each split is shown in Table \ref{tab:datasets}.

\paragraph{Generated Reviews EN-TH \cite{lowphansirikul2020scb}}
A dataset of 174,530 reviews generated by CTRL \cite{keskar2019ctrl} originally in English and then translated into Thai using Google Translate API. Each review is labeled with a rating from 1 to 5 stars. We only use the translated Thai reviews as a feature to predict the labels.

\paragraph{Prachathai67k \cite{prachathai67k}}
A dataset of 67,889 Thai news articles from Prachathai, a Thai online news site. Each article is labeled with a set of tags representing the themes of its content. Each article can have more than 1 tag. Possible tags are \textit{politics}, \textit{human\_rights}, \textit{quality\_of\_life}, \textit{international}, \textit{social}, \textit{environment}, \textit{economics}, \textit{culture}, \textit{labor}, \textit{national\_security}, \textit{ict}, and \textit{education}. We only use each article's headline as a feature to predict the tags.

\paragraph{ThaiNER \cite{WannaphongPhatthiyaphaibun_2019}}
A dataset of 6,348 Thai sentences that have been tokenized and annotated with NER (IOB format) and POS tags. Possible NER tags are \textit{DATE}, \textit{EMAIL}, \textit{LAW}, \textit{LEN}, \textit{LOCATION}, \textit{MONEY}, \textit{ORGANIZATION}, \textit{PERCENT}, \textit{PERSON}, \textit{PHONE}, \textit{TIME}, \textit{URL}, and \textit{ZIP}. Originally, the dataset also has the unconfirmed tag that represents tokens recognized by annotators as part of a named entity but whose precise tags are undecided. We thus replace every occurrence with O, reducing the number of possible labels by 1.

\paragraph{LST20 \cite{boonkwan2020annotation}}
A dataset of 74,180 Thai sentences that have been tokenized and annotated with NER (IOBE format) and POS tags. Possible NER tags are \textit{BRN}, \textit{DES}, \textit{DTM}, \textit{LOC}, \textit{MEA}, \textit{NUM}, \textit{ORG}, \textit{PER}, \textit{TRM}, and \textit{TTL} while possible POS tags are \textit{AJ}, \textit{AV}, \textit{AX}, \textit{CC}, \textit{CL}, \textit{FX}, \textit{IJ}, \textit{NG}, \textit{NN}, \textit{NU}, \textit{PA}, \textit{PR}, \textit{PS}, \textit{PU}, \textit{VV}, and \textit{XX}.

\paragraph{Thai-NNER \cite{buaphet-etal-2022-thai}}
A dataset of 4,894 Thai documents that have been tokenized and annotated with NER tags in the IOBES format. The annotation is done in a nested manner, with NER tags separated across 8 layers. The span of each named entity in a deeper layer will always be contained within the span of a named entity in shallower layers. We only fine-tune the models to predict NER tags on the shallowest layer. There are 104 possible NER tags, which are listed at the end of the dataset's associated paper.

\subsubsection{Benchmarking Models}

In addition to our model, PhayaThaiBERT, we also fine-tune other similar models for comparison, namely mBERT (bert-base-multilingual-cased) \cite{devlin2019bert}, XLM-R (xlm-roberta-base) \cite{conneau-etal-2020-unsupervised}, and WangchanBERTa (wangchanberta-base-att-spm-uncased) \cite{lowphansirikul2021wangchanberta}, all of which are pretrained transformer-based masked language models.

\subsubsection{Fine-tuning Procedure}

Our fine-tuning procedure is designed to be mostly similar to the one described in the Downstream Tasks section of the WangchanBERTa paper \cite{lowphansirikul2021wangchanberta}.

For WangchanBERTa and PhayaThaiBERT, the datasets are preprocessed in the same manner as described in the WangchanBERTa paper. For mBERT and XLM-R, the datasets are simply tokenized using their respective tokenizers. Although the datasets for token classification task already consist of pre-tokenized sentences, they are not tokenized in a way that can be directly used as input to the models. Each token is still required to be tokenized by each model's specific tokenizer, after which each token in the original datasets could be separated into a sequence of more than 1 token, in which case we decide to label only the first token, meaning the models will be trained to predict only the first token of each sequence.

For mBERT and XLM-R, we limit the length of the input to 512 tokens, the maximum allowed by the models. For WangchanBERTa and PhayaThaiBERT, we fine-tune each model 2 times, once with 416 as the length limit, the same value used during training, and once with 510 as the length limit, the maximum allowed by the models. The models with better performance are then selected and their performance is reported for each model. For token classification task, where every token needs to be accounted for, we simply split each example into chunks, each no longer than the length limit.

Table \ref{tab:fine_tuning_hyperparameters} shows the hyperparameters we use to fine-tune the models for each task. All fine-tuning is done using the Trainer API of Hugging Face's \textit{transformers} library \cite{wolf-etal-2020-transformers}.

\begin{table*}
    \footnotesize
    \centering
    \begin{tabular}{|l|c|c|c|}
        \hline
        Hyperparameter & \begin{tabular}{c}Single-label\\sequence\\classification\end{tabular} & \begin{tabular}{c}Multi-label\\sequence\\classification\end{tabular} & \begin{tabular}{c}Single-label\\token\\classification\end{tabular} \\
        \hline
        Peak learning rate & \multicolumn{3}{c|}{3e-5} \\
        \hline
        Warmup step ratio & \multicolumn{3}{c|}{0.1} \\
        \hline
        Weight decay & \multicolumn{3}{c|}{0.01} \\
        \hline
        Adam $\epsilon$ & \multicolumn{3}{c|}{1e-8} \\
        \hline
        Adam $\beta_1$ & \multicolumn{3}{c|}{0.9} \\
        \hline
        Adam $\beta_2$ & \multicolumn{3}{c|}{0.999} \\
        \hline
        FP16 & \multicolumn{3}{c|}{True} \\
        \hline
        Batch size & \multicolumn{2}{c|}{16} & 32 \\
        \hline
        Number of training epochs & \multicolumn{2}{c|}{3} & 6\textsuperscript{*} \\
        \hline
        Checkpointing steps & \multicolumn{3}{c|}{100\textsuperscript{**}} \\
        \hline
        Metric for best checkpoint & micro-average F1 & macro-average F1 & loss \\
        \hline
        \multicolumn{4}{c}{} \\
        \multicolumn{4}{l}{\begin{tabular}{l}* Except for Thai-NNER, which we train the models for 20 epochs to compensate for\\the fact that Thai-NNER has an extremely large tag set of over 400 unique NER tags\end{tabular}} \\
        \multicolumn{4}{l}{\begin{tabular}{l}** Except for ThaiNER and Thai-NNER, which we save a checkpoint of the models\\every 20 steps since these datasets have a very small training set\end{tabular}}
    \end{tabular}
    \caption{Hyperparameters used in the fine-tuning of each task.}
    \label{tab:fine_tuning_hyperparameters}
\end{table*}

\subsubsection{Metrics}

The metrics we use to evaluate the fine-tuned models are micro-average F1 score and macro-average F1 score. In case of NER task, the metrics are computed using seqeval's \textit{classification\_report} function \cite{seqeval}. In case of other tasks, they are computed using scikit-learn's \textit{classification\_report} function \cite{scikit-learn}.

\section{Results and Discussion}

\subsection{Pretraining}

The Validation loss computed during training reaches an apparent minimum at step 51,500 (epoch 2.3), after which it fails to decrease further even after another 3,800 update steps. We thus take the model's checkpoint at step 51,500 as our final pretrained model. The Validation loss at step 51,500 is 1.6524.

\subsection{Fine-tuning on downstream tasks}
\label{sec:fine_tuning_result}

\begin{table*}[ht]
    \footnotesize
    \centering
    \begin{tabular}{lcccc}
        \hline
        Dataset & mBERT & XLM-R & WangchanBERTa & PhayaThaiBERT \\
        \hline
        1. wisesight\_sentiment & 70.57 / 55.62 & 71.77 / 58.29 & 74.35 / 65.23 & \textbf{76.15} / \textbf{66.80} \\
        2. wongnai\_reviews & 58.08 / 38.67 & 62.73 / 51.18 & 63.86 / \textbf{53.26} & \textbf{64.02} / 52.74 \\
        3. yelp\_review\_full & 63.52 / 63.23 & \textbf{65.19} / \textbf{64.80} & 54.97 / 54.40 & 61.69 / 61.26 \\
        4. generated\_reviews\_enth & 61.79 / 56.04 & \textbf{65.06} / \textbf{60.28} & 64.75 / 59.91 & 64.85 / 59.44 \\
        5. prachathai67k & 63.90 / 52.95 & 66.63 / 58.01 & 67.51 / 59.05 & \textbf{69.11} / \textbf{61.10} \\
        6. thainer (ner) & 79.58 / 69.87 & 84.95 / 72.20 & 84.64 / 68.19 & \textbf{86.42} / \textbf{74.77} \\
        7. lst20 (pos) & 95.80 / 83.95 & 95.99 / 85.09 & 96.74 / \textbf{86.59} & \textbf{96.79} / 86.26 \\
        8. lst20 (ner) & 76.48 / 70.27 & \textbf{78.37} / 72.82 & 77.99 / \textbf{72.86} & 78.11 / 72.69 \\
        9. thai\_nner (layer 1) & 61.19 / 23.45 & 63.88 / 23.28 & 59.31 / 22.85 & \textbf{64.26} / \textbf{25.70} \\
        \hline
    \end{tabular}
    \caption{Fine-tuning results. The reported metrics are micro-average and macro-average F1 score respectively.}
    \label{tab:finetune_result}
\end{table*}

\begin{table*}
    \footnotesize
    \centering
    \begin{tabular}{lccc}
        \hline
        Dataset & mBERT & XLM-R & WangchanBERTa \\
        \hline
        1. wisesight\_sentiment & 70.05 / 57.81 & 73.57 / 62.21 & \textbf{76.19} / \textbf{67.05} \\
        2. wongnai\_reviews & 47.99 / 12.97 & 62.57 / \textbf{52.75} & \textbf{63.05} / 52.19 \\
        3. yelp\_review\_full & - & - & - \\
        4. generated\_reviews\_enth & 62.14 / 57.20 & \textbf{64.91} / \textbf{60.29} & 64.66 / 59.54 \\
        5. prachathai67k & 66.47 / 60.11 & 68.18 / 63.14 & \textbf{69.78} / \textbf{64.90} \\
        6. thainer (ner) & 81.48 / 73.97 & 83.25 / 67.23 & \textbf{86.49} / \textbf{79.29} \\
        7. lst20 (pos) & 96.44 / \textbf{85.86} & 96.57 / 85.00 & \textbf{96.62} / 85.44  \\
        8. lst20 (ner) & 75.05 / 68.25 & 73.61 / 68.67 & \textbf{78.01} / \textbf{72.25} \\
        9. thai\_nner (layer 1) & - & - & - \\
        \hline
    \end{tabular}
    \caption{Original fine-tuning results as reported in the WangchanBERTa paper \cite{lowphansirikul2021wangchanberta}.}
    \label{tab:original_result}
\end{table*}

The evaluation results of all fine-tuned models on test splits can be seen in Table \ref{tab:finetune_result}. The results originally reported in the WangchanBERTa paper is also given in Table \ref{tab:original_result} for comparison. While we try to fine-tune our models using procedures as close as possible to those in the original paper, we fail to reproduce the same level of performance for many models. Nevertheless, our results should give a valid comparison since all our fine-tuned models are guaranteed to come from the exact same procedures and environments.

Our model outperforms all other models in both micro-average and macro-average F1 scores in 4 out of the 9 datasets (1, 5, 6, 9). Our model outperforms all other models only in micro-average F1 scores in 2 out of the remaining 5 datasets (2, 7), where our model is defeated in both cases by WangchanBERTa in macro-average F1 scores. Among the remaining 3 datasets, XLM-R achieves both the highest micro-average and macro-average F1 scores in 2 of them (3, 4). For the last remaining dataset (8) the highest micro-average F1 score is achieved by XLM-R while the highest macro-average F1 score is achieved by WangchanBERTa.

However, if we only compare between WangchanBERTa and our model, our model defeats WangchanBERTa in both metrics in 5 out of the 9 datasets (1, 3, 5, 6, 9), with our model defeating WangchanBERTa in 1 out of the 2 metrics in the remaining 4 datasets (2, 4, 7, 8). In no case does WangchanBERTa defeat our model in both metrics. These results demonstrate that our model has indeed been improved beyond WangchanBERTa's capability, although at a cost of a significantly larger model size.

The outlying results in Yelp Review Full (3) and Generated Reviews EN-TH (4), where XLM-R achieved the highest score in both metrics, could be explained by the non-Thai-ness of their samples. Yelp Review Full consists entirely of English reviews while Generated Reviews EN-TH's examples are originally English texts that have been translated into Thai using a machine translation model. It could thus be reasonably expected that XLM-R, which has been trained extensively on English language pattern, would give better performance in such cases. Our results confirm the finding in the WangchanBERTa paper \cite{lowphansirikul2021wangchanberta}, in which XLM-R is also reported to give better performance for Generated Reviews EN-TH.

\subsection{Effect of Our Enhancement on OOV Rate}

Out-Of-Vocabulary (OOV) rate is the proportion of input texts that cannot be tokenized because there are no appropriate words in the tokenizer's vocabulary. In case of RoBERTa-based models, this is represented by <unk> tokens. Since every English alphabet is already part of WangchanBERTa's vocabulary, the OOV rate is already small even without our enhancement, with <unk> tokens appearing only in cases where unusual characters are used, such as emojis or words written in scripts other than Thai and basic Latin. Nevertheless, our enhancement manages to reduce this rate even further, as can be seen in Table \ref{tab:oov_rate}.

\begin{table*}[ht]
    \footnotesize
    \centering
    \begin{tabular}{lcccc}
        \hline
        \multirow{2}{*}{Dataset} & \multicolumn{2}{c}{WangchanBERTa} & \multicolumn{2}{c}{PhayaThaiBERT} \\
        & <unk> count & Percentage & <unk> count & Percentage \\
        \hline
        1. wisesight\_sentiment & 2,900 & 0.32\% & 34 & $\sim$0\% \\
        2. wongnai\_reviews & 3,855 & 0.05\% & 90 & $\sim$0\% \\
        3. yelp\_review\_full & 15 & $\sim$0\% & 0 & 0\% \\
        4. generated\_reviews\_enth & 99 & $\sim$0\% & 10 & $\sim$0\% \\
        5. prachathai67k & 241 & 0.02\% & 61 & $\sim$0\% \\
        6. thainer & 8 & $\sim$0\% & 1 & $\sim$0\% \\
        7. lst20 & 19 & $\sim$0\% & 15 & $\sim$0\% \\
        8. thai\_nner & 267 & 0.01\% & 23 & $\sim$0\% \\
        \hline
    \end{tabular}
    \caption{OOV rates of WangchanBERTa's tokenizer and our expanded tokenizer for each dataset.}
    \label{tab:oov_rate}
\end{table*}

\subsection{Do Unassimilated Loanwords Affect Models' Performance?}

Table \ref{tab:borrowing_rate} shows the proportion of samples that contain unassimilated English words in each dataset along with our model's performance gain compared to WangchanBERTa. We only show the statistics for English because English words are by far the most common loanwords in Thai texts.

For sequence classification datasets (1, 2, 3, 4, 5), there does not seem to be any relation between prevalence of unassimilated loanwords and performance except for Yelp Review Full (3). This is likely because the important pieces of text used to determine the labels are mostly in Thai, making any improvement to the understanding of foreign words unlikely to result in any significant performance gain. This of course does not apply to Yelp Review Full, which is entirely English, and it is here that we see a large gain due to our enhancement.

In constrast, for token classification datasets (6, 7, 8, 9), there seems to be a correlation between prevalence of unassimilated loanwords and performance gain. That is, the higher the prevalence of unassimilated English words, the higher the performance gain. The difference from previous datasets could be explained by the fact that token classification requires the models to understand every token in order to be able to assign their labels correctly. This means that any improvement to the understanding of foreign words should contribute directly to the accuracy of the models.

\begin{table*}[ht]
    \footnotesize
    \centering
    \begin{tabular}{lC{20ex}C{20ex}}
        \hline
        Dataset & Proportion of Samples with Unassimilated Loanwords & Performance Gain Compared to WangchanBERTa \\
        \hline
        1. wisesight\_sentiment & 27.59\% & +1.8 / +1.57 \\
        2. wongnai\_reviews & 37.20\% & +0.16 / -0.52 \\
        3. yelp\_review\_full & Entirely English & +6.72 / +9.86 \\
        4. generated\_reviews\_enth & 36.68\% & +0.1 / -0.47 \\
        5. prachathai67k & 8.54\% & +1.6 / +2.05 \\
        6. thainer & 10.90\% & +1.78 / +6.58 \\
        7. lst20 (pos) & 2.58\% & +0.05 / -0.33 \\
        8. lst20 (ner) & 2.58\% & +0.12 / -0.17 \\
        9. thai\_nner & 38.54\% & +4.95 / +2.85 \\
        \hline
    \end{tabular}
    \caption{Proportion of samples with unassimilated English words compared with performance gain for each dataset.}
    \label{tab:borrowing_rate}
\end{table*}

\section{Conclusion}

In this study, we introduce PhayaThaiBERT, a pretrained Thai masked language model based on WangchanBERTa that has been enhanced to improve its understanding of orthographically unassimilated loanwords. With WangchanBERTa as foundation, we improve upon it by expanding its vocabulary and pretraining it further on new and larger training data. Some modifications and techniques are applied to prevent loss of performance due to these procedures. We fine-tune the pretrained model on several tasks and datasets. The results show that our model has improved beyond WangchanBERTa's capability, albeit not decisive in some context. As with WangchanBERTa, our model also fails to achieve better performance than XLM-R for Generated Reviews EN-TH, a dataset whose texts are translated from English using a machine translation model.

\section*{Limitations}

Our model, while achieving better performance, is significantly larger than WangchanBERTa and thus will require significantly more computing resources to use. The sole source of the increase in model size is the added vocabulary, which contains a large number of foreign words that would not all be needed for every downstream application. Steps that could be taken to alleviate this problem might include model distillation \cite{sanh2020distilbert}, and exclusion of unnecessary vocabulary from the model.

Moreover, our training procedure is far from ideal, with many aspects based on arbitrary and subjective decision. While we are successful in defeating WangchanBERTa, performing more experiments to identify the optimal procedure and hyperparameters could potentially improve our model further. This has, however, been impossible given the time and resources available to us.

%\section*{Ethics Statement}

%That large language models tend to reproduce existing stereotypes and societal biases is a well-known fact. Our model is not immune to this tendency. For example, when asked to predict the masked token in the sentence "[Pronoun]\textthai{ทำงานเป็น}<mask>" (I work as a <mask>) our model assigns the highest score, 2 times higher than the next token, to the word "\textthai{แม่บ้าน}" (housewife/female cleaner) when the subject is referred to using the feminine pronoun "\textthai{ดิฉัน}" but does not do the same when the masculine pronoun "\textthai{ผม}" is used. While our model, a masked language model, is unlikely to cause any harm by itself, the inclusion of our model in larger systems could potentially amplify these stereotypes and biases. Any use of our model should thus be done with care not to cause any harm to any potential stakeholders.

%The training data we use is based entirely on publicly available data. This of course does not entail that every part of it will still be considered "public" information in the future. While it has been shown that generative pretrained language models could potentially be subjected to training data extraction attacks that cause them to reveal the original training data \cite{ishihara2023training}, to the best of our knowledge, the same has yet to be demonstrated for masked language models, which are not generative in nature. Therefore, the possibility of violation to the right to privacy should be minimal.

\section*{Acknowledgements}

We would like to thank Lalita Lowphansirikul, one of the authors of the WangchanBERTa paper, for providing us with various pieces of advice and insights about the training of large masked language models. We thank the WangchanBERTa team for providing us with the original data used in the training of WangchanBERTa. We thank NSTDA Supercomputer Center for giving us free of charge access to the LANTA HPC cluster (project lt200063). Finally, we would also like to thank our colleagues at ClickNext who contribute in various ways to this study, especially Jakkrit Diloksrisakul, Pontakorn Trakuekul, Jiratchaya Sutthirod, and Tanhaporn Tosirikul.

\bibliography{anthology,custom}

\begin{thebibliography}{29}
\expandafter\ifx\csname natexlab\endcsname\relax\def\natexlab#1{#1}\fi

\bibitem[{Boonkwan et~al.(2020)Boonkwan, Luantangsrisuk, Phaholphinyo, Kriengket, Leenoi, Phrombut, Boriboon, Kosawat, and Supnithi}]{boonkwan2020annotation}
Prachya Boonkwan, Vorapon Luantangsrisuk, Sitthaa Phaholphinyo, Kanyanat Kriengket, Dhanon Leenoi, Charun Phrombut, Monthika Boriboon, Krit Kosawat, and Thepchai Supnithi. 2020.
\newblock \href {http://arxiv.org/abs/2008.05055} {The annotation guideline of lst20 corpus}.

\bibitem[{Buaphet et~al.(2022)Buaphet, Udomcharoenchaikit, Limkonchotiwat, Rutherford, and Nutanong}]{buaphet-etal-2022-thai}
Weerayut Buaphet, Can Udomcharoenchaikit, Peerat Limkonchotiwat, Attapol Rutherford, and Sarana Nutanong. 2022.
\newblock \href {https://doi.org/10.18653/v1/2022.findings-acl.116} {{T}hai nested named entity recognition corpus}.
\newblock In \emph{Findings of the Association for Computational Linguistics: ACL 2022}, pages 1473--1486, Dublin, Ireland. Association for Computational Linguistics.

\bibitem[{Chotirat et~al.(2022)Chotirat, Meesad, and Unger}]{thai_question_classification}
Saranlita Chotirat, Phayung Meesad, and Herwig Unger. 2022.
\newblock \href {https://doi.org/10.1109/RI2C56397.2022.9910313} {Question classification from thai sentences by considering word context to question generation}.
\newblock In \emph{2022 Research, Invention, and Innovation Congress: Innovative Electricals and Electronics (RI2C)}, pages 9--14.

\bibitem[{Conneau et~al.(2020)Conneau, Khandelwal, Goyal, Chaudhary, Wenzek, Guzm{\'a}n, Grave, Ott, Zettlemoyer, and Stoyanov}]{conneau-etal-2020-unsupervised}
Alexis Conneau, Kartikay Khandelwal, Naman Goyal, Vishrav Chaudhary, Guillaume Wenzek, Francisco Guzm{\'a}n, Edouard Grave, Myle Ott, Luke Zettlemoyer, and Veselin Stoyanov. 2020.
\newblock \href {https://doi.org/10.18653/v1/2020.acl-main.747} {Unsupervised cross-lingual representation learning at scale}.
\newblock In \emph{Proceedings of the 58th Annual Meeting of the Association for Computational Linguistics}, pages 8440--8451, Online. Association for Computational Linguistics.

\bibitem[{cstorm125 and lukkiddd(2019)}]{prachathai67k}
cstorm125 and lukkiddd. 2019.
\newblock prachathai67k.
\newblock \url{https://github.com/PyThaiNLP/prachathai-67k}.

\bibitem[{Devlin et~al.(2019)Devlin, Chang, Lee, and Toutanova}]{devlin2019bert}
Jacob Devlin, Ming-Wei Chang, Kenton Lee, and Kristina Toutanova. 2019.
\newblock \href {http://arxiv.org/abs/1810.04805} {Bert: Pre-training of deep bidirectional transformers for language understanding}.

\bibitem[{Gugger et~al.(2022)Gugger, Debut, Wolf, Schmid, Mueller, Mangrulkar, Sun, and Bossan}]{accelerate}
Sylvain Gugger, Lysandre Debut, Thomas Wolf, Philipp Schmid, Zachary Mueller, Sourab Mangrulkar, Marc Sun, and Benjamin Bossan. 2022.
\newblock Accelerate: Training and inference at scale made simple, efficient and adaptable.
\newblock \url{https://github.com/huggingface/accelerate}.

\bibitem[{Howard and Ruder(2018)}]{howard-ruder-2018-universal}
Jeremy Howard and Sebastian Ruder. 2018.
\newblock \href {https://doi.org/10.18653/v1/P18-1031} {Universal language model fine-tuning for text classification}.
\newblock In \emph{Proceedings of the 56th Annual Meeting of the Association for Computational Linguistics (Volume 1: Long Papers)}, pages 328--339, Melbourne, Australia. Association for Computational Linguistics.

\bibitem[{Keskar et~al.(2019)Keskar, McCann, Varshney, Xiong, and Socher}]{keskar2019ctrl}
Nitish~Shirish Keskar, Bryan McCann, Lav~R. Varshney, Caiming Xiong, and Richard Socher. 2019.
\newblock \href {http://arxiv.org/abs/1909.05858} {Ctrl: A conditional transformer language model for controllable generation}.

\bibitem[{Khamphakdee and Seresangtakul(2023)}]{thai_sentiment_analysis}
Nattawat Khamphakdee and Pusadee Seresangtakul. 2023.
\newblock \href {https://doi.org/10.3390/data8050090} {An efficient deep learning for thai sentiment analysis}.
\newblock \emph{Data}, 8(5).

\bibitem[{Kudo(2018)}]{kudo-2018-subword}
Taku Kudo. 2018.
\newblock \href {https://doi.org/10.18653/v1/P18-1007} {Subword regularization: Improving neural network translation models with multiple subword candidates}.
\newblock In \emph{Proceedings of the 56th Annual Meeting of the Association for Computational Linguistics (Volume 1: Long Papers)}, pages 66--75, Melbourne, Australia. Association for Computational Linguistics.

\bibitem[{Kudo and Richardson(2018)}]{kudo-richardson-2018-sentencepiece}
Taku Kudo and John Richardson. 2018.
\newblock \href {https://doi.org/10.18653/v1/D18-2012} {{S}entence{P}iece: A simple and language independent subword tokenizer and detokenizer for neural text processing}.
\newblock In \emph{Proceedings of the 2018 Conference on Empirical Methods in Natural Language Processing: System Demonstrations}, pages 66--71, Brussels, Belgium. Association for Computational Linguistics.

\bibitem[{Lhoest et~al.(2021)Lhoest, Villanova~del Moral, Jernite, Thakur, von Platen, Patil, Chaumond, Drame, Plu, Tunstall, Davison, {\v{S}}a{\v{s}}ko, Chhablani, Malik, Brandeis, Le~Scao, Sanh, Xu, Patry, McMillan-Major, Schmid, Gugger, Delangue, Matussi{\`e}re, Debut, Bekman, Cistac, Goehringer, Mustar, Lagunas, Rush, and Wolf}]{lhoest-etal-2021-datasets}
Quentin Lhoest, Albert Villanova~del Moral, Yacine Jernite, Abhishek Thakur, Patrick von Platen, Suraj Patil, Julien Chaumond, Mariama Drame, Julien Plu, Lewis Tunstall, Joe Davison, Mario {\v{S}}a{\v{s}}ko, Gunjan Chhablani, Bhavitvya Malik, Simon Brandeis, Teven Le~Scao, Victor Sanh, Canwen Xu, Nicolas Patry, Angelina McMillan-Major, Philipp Schmid, Sylvain Gugger, Cl{\'e}ment Delangue, Th{\'e}o Matussi{\`e}re, Lysandre Debut, Stas Bekman, Pierric Cistac, Thibault Goehringer, Victor Mustar, Fran{\c{c}}ois Lagunas, Alexander Rush, and Thomas Wolf. 2021.
\newblock \href {https://doi.org/10.18653/v1/2021.emnlp-demo.21} {Datasets: A community library for natural language processing}.
\newblock In \emph{Proceedings of the 2021 Conference on Empirical Methods in Natural Language Processing: System Demonstrations}, pages 175--184, Online and Punta Cana, Dominican Republic. Association for Computational Linguistics.

\bibitem[{Liu et~al.(2019)Liu, Ott, Goyal, Du, Joshi, Chen, Levy, Lewis, Zettlemoyer, and Stoyanov}]{liu2019roberta}
Yinhan Liu, Myle Ott, Naman Goyal, Jingfei Du, Mandar Joshi, Danqi Chen, Omer Levy, Mike Lewis, Luke Zettlemoyer, and Veselin Stoyanov. 2019.
\newblock \href {http://arxiv.org/abs/1907.11692} {Roberta: A robustly optimized bert pretraining approach}.

\bibitem[{Lowphansirikul et~al.(2021{\natexlab{a}})Lowphansirikul, Polpanumas, Jantrakulchai, and Nutanong}]{lowphansirikul2021wangchanberta}
Lalita Lowphansirikul, Charin Polpanumas, Nawat Jantrakulchai, and Sarana Nutanong. 2021{\natexlab{a}}.
\newblock \href {http://arxiv.org/abs/2101.09635} {Wangchanberta: Pretraining transformer-based thai language models}.

\bibitem[{Lowphansirikul et~al.(2021{\natexlab{b}})Lowphansirikul, Polpanumas, Rutherford, and Nutanong}]{lowphansirikul2020scb}
Lalita Lowphansirikul, Charin Polpanumas, Attapol~T. Rutherford, and Sarana Nutanong. 2021{\natexlab{b}}.
\newblock \href {https://doi.org/10.1007/s10579-021-09536-6} {A large english{\textendash}thai parallel corpus from the web and machine-generated text}.
\newblock \emph{Language Resources and Evaluation}, 56(2):477--499.

\bibitem[{Martin et~al.(2020)Martin, Muller, Ortiz~Su{\'a}rez, Dupont, Romary, de~la Clergerie, Seddah, and Sagot}]{martin-etal-2020-camembert}
Louis Martin, Benjamin Muller, Pedro~Javier Ortiz~Su{\'a}rez, Yoann Dupont, Laurent Romary, {\'E}ric de~la Clergerie, Djam{\'e} Seddah, and Beno{\^\i}t Sagot. 2020.
\newblock \href {https://doi.org/10.18653/v1/2020.acl-main.645} {{C}amem{BERT}: a tasty {F}rench language model}.
\newblock In \emph{Proceedings of the 58th Annual Meeting of the Association for Computational Linguistics}, pages 7203--7219, Online. Association for Computational Linguistics.

\bibitem[{Nakayama(2018)}]{seqeval}
Hiroki Nakayama. 2018.
\newblock \href {https://github.com/chakki-works/seqeval} {{seqeval}: A python framework for sequence labeling evaluation}.
\newblock Software available from https://github.com/chakki-works/seqeval.

\bibitem[{Nguyen and Tuan~Nguyen(2020)}]{nguyen-tuan-nguyen-2020-phobert}
Dat~Quoc Nguyen and Anh Tuan~Nguyen. 2020.
\newblock \href {https://doi.org/10.18653/v1/2020.findings-emnlp.92} {{P}ho{BERT}: Pre-trained language models for {V}ietnamese}.
\newblock In \emph{Findings of the Association for Computational Linguistics: EMNLP 2020}, pages 1037--1042, Online. Association for Computational Linguistics.

\bibitem[{Noiyoo and Thutkawkornpin(2023)}]{thai_essay_checking}
Nichaphan Noiyoo and Jessada Thutkawkornpin. 2023.
\newblock \href {https://doi.org/10.1109/JCSSE58229.2023.10201941} {A comparison of machine learning and neural network algorithms for an automated thai essay quality checking}.
\newblock In \emph{2023 20th International Joint Conference on Computer Science and Software Engineering (JCSSE)}, pages 482--487.

\bibitem[{Pankam et~al.(2023)Pankam, Limkonchotiwat, and Chuangsuwanich}]{thai_misspelling_correction}
Idhibhat Pankam, Peerat Limkonchotiwat, and Ekapol Chuangsuwanich. 2023.
\newblock \href {https://doi.org/10.1109/JCSSE58229.2023.10202006} {Two-stage thai misspelling correction based on pre-trained language models}.
\newblock In \emph{2023 20th International Joint Conference on Computer Science and Software Engineering (JCSSE)}, pages 7--12.

\bibitem[{Pedregosa et~al.(2011)Pedregosa, Varoquaux, Gramfort, Michel, Thirion, Grisel, Blondel, Prettenhofer, Weiss, Dubourg, Vanderplas, Passos, Cournapeau, Brucher, Perrot, and {{\'E}}douard Duchesnay}]{scikit-learn}
Fabian Pedregosa, Ga{{\"e}}l Varoquaux, Alexandre Gramfort, Vincent Michel, Bertrand Thirion, Olivier Grisel, Mathieu Blondel, Peter Prettenhofer, Ron Weiss, Vincent Dubourg, Jake Vanderplas, Alexandre Passos, David Cournapeau, Matthieu Brucher, Matthieu Perrot, and {{\'E}}douard Duchesnay. 2011.
\newblock \href {http://jmlr.org/papers/v12/pedregosa11a.html} {Scikit-learn: Machine learning in python}.
\newblock \emph{Journal of Machine Learning Research}, 12(85):2825--2830.

\bibitem[{Phatthiyaphaibun(2019)}]{WannaphongPhatthiyaphaibun_2019}
Wannaphong Phatthiyaphaibun. 2019.
\newblock \href {https://doi.org/10.5281/ZENODO.3550546} {wannaphongcom/thai-ner: Thainer 1.3}.

\bibitem[{Sanh et~al.(2020)Sanh, Debut, Chaumond, and Wolf}]{sanh2020distilbert}
Victor Sanh, Lysandre Debut, Julien Chaumond, and Thomas Wolf. 2020.
\newblock \href {http://arxiv.org/abs/1910.01108} {Distilbert, a distilled version of bert: smaller, faster, cheaper and lighter}.

\bibitem[{Suriyawongkul et~al.(2019)Suriyawongkul, Chuangsuwanich, Chormai, and Polpanumas}]{bact_2019_3457447}
Arthit Suriyawongkul, Ekapol Chuangsuwanich, Pattarawat Chormai, and Charin Polpanumas. 2019.
\newblock \href {https://doi.org/10.5281/zenodo.3457447} {Pythainlp/wisesight-sentiment: First release}.

\bibitem[{Wolf et~al.(2020)Wolf, Debut, Sanh, Chaumond, Delangue, Moi, Cistac, Rault, Louf, Funtowicz, Davison, Shleifer, von Platen, Ma, Jernite, Plu, Xu, Le~Scao, Gugger, Drame, Lhoest, and Rush}]{wolf-etal-2020-transformers}
Thomas Wolf, Lysandre Debut, Victor Sanh, Julien Chaumond, Clement Delangue, Anthony Moi, Pierric Cistac, Tim Rault, Remi Louf, Morgan Funtowicz, Joe Davison, Sam Shleifer, Patrick von Platen, Clara Ma, Yacine Jernite, Julien Plu, Canwen Xu, Teven Le~Scao, Sylvain Gugger, Mariama Drame, Quentin Lhoest, and Alexander Rush. 2020.
\newblock \href {https://doi.org/10.18653/v1/2020.emnlp-demos.6} {Transformers: State-of-the-art natural language processing}.
\newblock In \emph{Proceedings of the 2020 Conference on Empirical Methods in Natural Language Processing: System Demonstrations}, pages 38--45, Online. Association for Computational Linguistics.

\bibitem[{Wongnai.com(2018)}]{wongnai_reviews}
Wongnai.com. 2018.
\newblock Wongnai-corpus.
\newblock \url{https://github.com/wongnai/wongnai-corpus}.

\bibitem[{Wu et~al.(2016)Wu, Schuster, Chen, Le, Norouzi, Macherey, Krikun, Cao, Gao, Macherey, Klingner, Shah, Johnson, Liu, Łukasz Kaiser, Gouws, Kato, Kudo, Kazawa, Stevens, Kurian, Patil, Wang, Young, Smith, Riesa, Rudnick, Vinyals, Corrado, Hughes, and Dean}]{wu2016googles}
Yonghui Wu, Mike Schuster, Zhifeng Chen, Quoc~V. Le, Mohammad Norouzi, Wolfgang Macherey, Maxim Krikun, Yuan Cao, Qin Gao, Klaus Macherey, Jeff Klingner, Apurva Shah, Melvin Johnson, Xiaobing Liu, Łukasz Kaiser, Stephan Gouws, Yoshikiyo Kato, Taku Kudo, Hideto Kazawa, Keith Stevens, George Kurian, Nishant Patil, Wei Wang, Cliff Young, Jason Smith, Jason Riesa, Alex Rudnick, Oriol Vinyals, Greg Corrado, Macduff Hughes, and Jeffrey Dean. 2016.
\newblock \href {http://arxiv.org/abs/1609.08144} {Google's neural machine translation system: Bridging the gap between human and machine translation}.

\bibitem[{Zhang et~al.(2015)Zhang, Zhao, and LeCun}]{NIPS2015_250cf8b5}
Xiang Zhang, Junbo Zhao, and Yann LeCun. 2015.
\newblock \href {https://proceedings.neurips.cc/paper_files/paper/2015/file/250cf8b51c773f3f8dc8b4be867a9a02-Paper.pdf} {Character-level convolutional networks for text classification}.
\newblock In \emph{Advances in Neural Information Processing Systems}, volume~28. Curran Associates, Inc.

\end{thebibliography}
\bibliographystyle{acl_natbib}

\appendix
\onecolumn

\section{Sources of Training Data}
Below is the list of sources of raw texts that we use to create our training data. For datasets that contain multiple languages, we only take their subsets that contain Thai. The listed data sizes are measured after preprocessing but before tokenization.

\vspace{0.5cm}

    \begin{tabularx}{\textwidth}{p{25ex}cX}
        \hline
        Dataset & Data size & Source \\
        \hline
        40-Thai-Children-Stories & 416KB & https://github.com/dsmlr/40-Thai-Children-Stories \\
        Thai QA & 3.2MB & https://aiforthai.in.th/corpus.php (registration required) \\
        BeneficialTweetCOVID-19 & 1.9MB & https://github.com/ENEmyr/BeneficialTweetCOVID-19 \\
        mr-tydi & 660KB & https://huggingface.co/datasets/castorini/mr-tydi \\
        CC100-Thai & 73GB & https://autonlp.ai/datasets/cc100-thai \\
        miracl-th-corpus-22-12 & 480MB & https://huggingface.co/datasets/Cohere/miracl-th-corpus-22-12 \\
        miracl-th-queries-22-12 & 5.6MB & https://huggingface.co/datasets/Cohere/miracl-th-queries-22-12 \\
        common\_voice & 1.4MB & https://huggingface.co/datasets/common\_voice \\
        xlsum & 9.2MB & https://huggingface.co/datasets/csebuetnlp/xlsum \\
        Thaisong-ML & 63MB & https://github.com/aofiee/Dataset-Thaisong-ML \\
        flores\_101 & 716KB & https://huggingface.co/datasets/gsarti/flores\_101 \\
        wiki\_lingua & 20MB & https://huggingface.co/datasets/GEM/wiki\_lingua \\
        generated\_reviews\_enth & 123MB & https://huggingface.co/datasets/generated\_reviews\_enth \\
        han-corf-dataset-v1.0 & 1.1MB & https://huggingface.co/datasets/pythainlp/han-corf-dataset-v1.0 \\
        xnli2.0\_thai & 1.1MB & https://huggingface.co/datasets/Harsit/xnli2.0\_thai \\
        hc3-24k-th & 88MB & https://huggingface.co/datasets/Thaweewat/hc3-24k-th \\
        helltaker\_thai\_localization & 72KB & https://github.com/nekopaldee/helltaker\_thai\_localization \\
        iapp\_wiki\_qa\_squad & 5.3MB & https://huggingface.co/datasets/iapp\_wiki\_qa\_squad \\
        tydiqa-goldp & 5MB & https://huggingface.co/datasets/khalidalt/tydiqa-goldp \\
        tydiqa-primary & 158MB & https://huggingface.co/datasets/khalidalt/tydiqa-primary \\
        krathu-500 & 55MB & https://github.com/Pittawat2542/krathu-500 \\
        LimeSoda & 4.2MB & https://github.com/byinth/LimeSoda \\
        Bactrian-X & 107MB & https://huggingface.co/datasets/MBZUAI/Bactrian-X \\
        mc4 & 41GB & https://huggingface.co/datasets/mc4 \\
        blognone-20230430 & 47MB & https://huggingface.co/datasets/Noxturnix/blognone-20230430 \\
        xsum\_eng2thai & 71MB & https://huggingface.co/datasets/potsawee/xsum\_eng2thai \\
        Prime Minister 29 & 96KB & https://github.com/PyThaiNLP/lexicon-thai/tree/master/thai-corpus/Prime\%20Minister\%2029 \\
        traditional-thai-songs & 836KB & https://github.com/jaypeerachai/python-search-engine-for-thai-songs \\
        sentiment\_analysis\_thai & 528KB & https://github.com/JagerV3/sentiment\_analysis\_thai \\
        Thai-Clickbait & 13MB & https://github.com/9meo/Thai-Clickbait \\

                thai-covid-19-situation & 13MB & https://github.com/PyThaiNLP/thai-covid-19-situation \\
        thaigov-v2-corpus & 267MB & https://github.com/PyThaiNLP/thaigov-v2-corpus \\
        thai-joke-corpus & 556KB & https://github.com/iapp-technology/thai-joke-corpus \\
        Thai-Lao-Parallel-Corpus & 628KB & https://github.com/PyThaiNLP/Thai-Lao-Parallel-Corpus \\        
        thailaw & 349MB & https://huggingface.co/datasets/pythainlp/thailaw \\
        thai-mscoco-2014-captions & 71MB & https://huggingface.co/datasets/patomp/thai-mscoco-2014-captions \\
        TNBT & 2.6MB & https://github.com/pepa65/TNBT \\
        Thai Plagiarism & 882MB & https://aiforthai.in.th/corpus.php (registration required) \\

        \hline
        
    \end{tabularx}

    \centering
    \begin{tabularx}{\textwidth}{p{25ex}cX}
        \hline
        Dataset & Data size & Source \\
        \hline
        ThaiQA\_LST20 & 13MB & https://huggingface.co/datasets/SuperAI2-Machima/ThaiQA\_LST20 \\
        thaiqa\_squad & 684KB & https://huggingface.co/datasets/thaiqa\_squad \\
        Thai-Semantic-Textual-Similarity-Benchmark & 280KB & https://huggingface.co/datasets/mrp/Thai-Semantic-Textual-Similarity-Benchmark \\
        Thai-sentence & 68KB & https://github.com/PyThaiNLP/Thai-sentence \\
        Thai\_sentiment & 2.3MB & https://huggingface.co/datasets/sepidmnorozy/Thai\_\allowbreak sentiment \\
        thai\_toxicity\_tweet & 496KB & https://huggingface.co/datasets/thai\_toxicity\_tweet \\
        Thai-UCC & 24MB & https://huggingface.co/datasets/nakcnx/Thai-UCC \\
        thai\_wikipedia (20230101) & 666MB & https://github.com/PyThaiNLP/ThaiWiki-clean \\
        Thai Wiki QA & 667MB & https://copycatch.in.th/corpus/thai-wikiqa-nsc2020.html \\
        alpaca-cleaned-52k-th & 89MB & https://huggingface.co/datasets/Thaweewat/alpaca-cleaned-52k-th \\
        alpaca-finance-43k-th & 64MB & https://huggingface.co/datasets/Thaweewat/alpaca-finance-43k-th \\
        chain-of-thought-74k-th & 56MB & https://huggingface.co/datasets/Thaweewat/chain-of-thought-74k-th \\
        chatmed-5k-th & 4.3MB & https://huggingface.co/datasets/Thaweewat/chatmed-5k-th \\
        codegen-th & 1.1MB & https://huggingface.co/datasets/Thaweewat/codegen-th \\
        databricks-dolly-15k-th & 27MB & https://huggingface.co/datasets/Thaweewat/databricks-dolly-15k-th \\
        gpteacher-20k-th & 24MB & https://huggingface.co/datasets/Thaweewat/gpteacher-20k-th \\
        hc3-24k-th & 108MB & https://huggingface.co/datasets/Thaweewat/hc3-24k-th \\
        HealthCareMagic-100k-th & 285MB & https://huggingface.co/datasets/Thaweewat/HealthCareMagic-100k-th \\
        hh-rlhf-th & 252MB & https://huggingface.co/datasets/Thaweewat/hh-rlhf-th \\
        instruction-wild-52k-th & 96MB & https://huggingface.co/datasets/Thaweewat/instruction-wild-52k-th \\
        LaMini-instruction-th & 2.5GB & https://huggingface.co/datasets/Thaweewat/LaMini-instruction-th \\
        onet-m6-social & 180KB & https://huggingface.co/datasets/Thaweewat/onet-m6-social \\
        pobpad & 191MB & https://huggingface.co/datasets/Thaweewat/pobpad \\
        universal\_dependencies & 292KB & https://huggingface.co/datasets/universal\_dependencies \\
        Sanook-news & 2.3MB & https://github.com/sorayutmild/Unsupervised-Thai-Document-Clustering-with-Sanook-news \\
        VISTEC-TP-TH-21 & 38MB & https://github.com/mrpeerat/OSKut/tree/main/VISTEC-TP-TH-2021 \\
        tatoeba.tha & 52KB & https://huggingface.co/datasets/xtreme/viewer/tatoeba.tha \\
        thai-instructions-rallio & 8.6MB & https://huggingface.co/datasets/yadapruk/thai-instructions-rallio \\
        OSCAR-2301 & 69GB & https://huggingface.co/datasets/oscar-corpus/OSCAR-2301 \\
            Thai News Articles & 255MB & https://thematter.co/\break https://themomentum.co/\break https://thestandard.co/ \\
    pantip-large & 34.2GB & Raw texts provided by the WangchanBERTa team. Originally packaged by \href{https://www.facebook.com/ChaosTheoryCompany/}{Chaos Theory}. \\
    wisesight-large & 62.5GB & Raw texts provided by the WangchanBERTa team. Originally packaged by \href{https://wisesight.com/}{Wisesight}. \\
        \hline
    \end{tabularx}

\section{Hyperparameters}
Below is the specifications of the best-performing model hyperparameters used in the training process of PhayaThaiBERT. * indicates hyperparameters that are involved in some irregularities during training. 

\vspace{1cm}

\centering
    \begin{tabular}{lc}
        \hline
        \textbf{Hyperparameter} & \textbf{Value} \\
        \hline
        Number of layers & 12 \\
        Hidden size & 768 \\
        FFN hidden size & 3,072 \\
        Attention heads & 12 \\
        Dropout & 0.1 \\
        Attention dropout & 0.1 \\
        Max sequence length & 416 \\
        Batch size per GPU & 8 \\
        Number of GPUs & 16 \\
        Gradient accumulation steps & 32 \\
        Effective batch size & 4,096 \\
        Warmup steps & 24,000 \\
        Peak learning rate & 3e-4 \\
        Learning rate decay & Linear* \\
        Max steps & 500,000 \\
        Weight decay & 0.01 \\
        Adam $\epsilon$ & 1e-6 \\
        Adam $\beta_1$ & 0.9 \\
        Adam $\beta_2$ & 0.999 \\
        FP16 & True, False* \\
        \hline
    \end{tabular}

\end{document}